\documentclass{article}

\usepackage{arxiv}

\usepackage[utf8]{inputenc} 
\usepackage[T1]{fontenc}    
\usepackage{hyperref}       
\usepackage{url}            
\usepackage{booktabs}       
\usepackage{amsfonts}       
\usepackage{nicefrac}       
\usepackage{microtype}      
\usepackage{lipsum}		
\usepackage{graphicx}
\usepackage{biblatex} 
\usepackage{doi}
\usepackage{amsmath}
\usepackage{todonotes}
\usepackage{mdframed}
\usepackage{float}
\usepackage[export]{adjustbox}
\usepackage{subcaption}

\usepackage{amsthm}

\theoremstyle{definition}
\newtheorem{definition}{Definition}[section]

\newtheorem{theorem}{Theorem}[section]

\title{Rotation Equivariant Operators for Machine Learning on Scalar and Vector Fields}

\author{ Paul Shen\\
	Carnegie Mellon University\\
	\texttt{xingpins@andrew.cmu.edu} 
	\And
	Michael F. Herbst \\
	 RWTH Aachen University\\
	\texttt{herbst@acom.rwth-aachen.de} 
	\And
	Venkat Viswanathan \\
	Carnegie Mellon University \\
	\texttt{venkvis@cmu.edu} 
}

\hypersetup{
pdftitle={},
pdfsubject={q-bio.NC, q-bio.QM},
pdfauthor={David S.~Hippocampus, Elias D.~Striatum},
pdfkeywords={First keyword, Second keyword, More},
}

\begin{document}
\maketitle

\begin{abstract}
	We develop theory and software for rotation equivariant operators on scalar and vector fields, with diverse applications in simulation, optimization and machine learning. Rotation equivariance (covariance) means all fields in the system rotate together, implying spatially invariant dynamics that preserve symmetry. Extending the convolution theorems of linear time invariant systems, we theorize that linear equivariant operators are characterized by tensor field convolutions using an appropriate product between the input field and a radially symmetric kernel field. Most Green's functions and differential operators are in fact equivariant operators, which can also fit unknown symmetry preserving dynamics by parameterizing the radial function. We implement the Julia package EquivariantOperators.jl for fully differentiable finite difference equivariant operators on scalar, vector and higher order tensor fields in 2d/3d. It can run forwards for simulation or image processing, or be back propagated for computer vision, inverse problems and optimal control. Code at \url{https://aced-differentiate.github.io/EquivariantOperators.jl/}
\end{abstract}

\keywords{Rotation equivariant neural networks \and Symmetry preserving machine learning \and \and Convolutional neural networks \and Tensor fields \and Scalar fields \and Vector fields \and Fourier neural operators \and Neural PDE \and Physics informed neural networks \and inverse problems \and parameter estimation}

\section{Introduction}
Scalar and vector fields are ubiquitous in science and engineering. In machine learning, vector field components are treated as images to which convolutional neural networks (CNNs) are applied. This results in big models, slow training and poor generalization. This fails to account for rotational equivariance inherent in physical systems. Physical laws are isotropic. Rotating the input should rotate the output the same way thus preserving symmetry.

How can we enforce rotation equivariance in CNNs? Easy! Simply make the filters radially symmetric \cite{thomas}. This has the added benefit of parameterizing a 2d/3d kernel by a 1d scalar radial function. Equivariant convolutions respects symmetries and generalizes robustly, while having magnitudes fewer parameters and training in fraction of the time. This is unsurprising as equivariance powerfully constrains the model search space.

Applications go beyond machine learning. Physical laws using differential operators and field interactions are naturally equivariant. As such, equivariant operators also qualify for finite difference time domain simulation of partial differential equations (PDEs), as well as parameter estimation, inverse problems and optimal control. Thus we forumulate equivariant operators as extensions of linear systems theory.

From linear systems theory, the output $\mathbf{v}$ of any linear translation invariant (LTI) system can be written as the convolution of the input $\mathbf{u}$ and a characteristic impulse response $\mathbf{h}$ \cite{chen}. The coordinate needs not be time but can also be space, allowing extension of the theory to scalar fields in 2d/3d. We now contribute 2 extensions. First, we can generalize beyond scalar fields to vector fields and higher order tensor fields by modifying the product used in convolution. Scalar vector product, dot product, cross product and other tensor product can replace the scalar product depending on the field type of the input, filter and output. Second, equivariance forces the filter field to be radially symmetric, eg separable into a scalar radial function and the unit scalar or vector field.

\section{Theory}
\subsection{Linear equivariant operators}
We work with operators on tensor fields including scalar and vector fields. A "tensor" means different things to different people. Here, tensor is an object to which rotations can be applied. A tensor field is then a mapping from 2d/3d to a tensor. The simplest tensor is a scalar, which is actually invariant to rotations and by definition has rotation order $l=0$. Next is a vector, which has $l=1$. Higher order tensors can be represented by circular harmonics in 2d and spherical harmonics in 3d 

\begin{equation}
    l=
    \begin{cases}
    0 & \text{for scalar fields}\\
    1 & \text{for vector fields}\\
    >1 & \text{for higher order 2d circular harmonics or 3d spherical harmonics fields}\\
    \end{cases}
\end{equation}

\begin{mdframed}
\begin{definition}[Linear equivariant operators]
A Linear equivariant operator maps a tensor field (eg scalar or vector field) to another tensor field not necessarily of the same rotation order $l$ while satisfying:
\begin{enumerate}
  \item Linearity: $L(au+bv)=aL(u)+bL(v)$
  \item Translation invariance: $L \circ \text{trans}=\text{trans} \circ L$ $\forall$ translations
  \item Rotation equivariance: $L \circ \text{rot}=\text{rot} \circ L$ $\forall$ rotations
\end{enumerate}
\end{definition}
\end{mdframed}

\begin{mdframed}
\begin{definition}[Tensor field convolution]
Tensor field convolutions extend convolutions by augmenting the scalar product with a tensor product $\otimes$ relating the field types of inputs and output.

\begin{equation}
\begin{split}
\mathbf{u} \ast \mathbf{h}
& = \int \limits _{\mathbb{R}^n} \mathbf{u}(\mathbf{\tilde{r}}) \otimes
\mathbf{h}(\mathbf{r}-\mathbf{\tilde{r}}) d\mathbf{\tilde{r}}\\
\end{split}
\end{equation}

Possible choices for $\otimes$ include
\begin{equation}
    \mathbf{v}=\mathbf{u} \otimes  \mathbf{h}=
    \begin{cases}
u \mathbf{h} &      \text{scalar product for $l_u=0$ and $l_h=l_v$}\\
 \mathbf{u} \cdot \mathbf{h} & \text{dot product for $l_v=0$ and $l_u=l_h$}\\
\mathbf{u} \times \mathbf{h} &  \text{cross product for $l_u=l_h=l_v=1$}\\
    \end{cases}
    \end{equation}
    \end{definition}
\end{mdframed}

\begin{mdframed}
\begin{theorem}[Convolutional theorem for equivariant operators]
\label{pythagorean}
Let $L$ be a linear equivariant operator iff the following:
\begin{enumerate}
    \item Applying $L$ amounts to a tensor field convolution between the input field and a filter kernel field $\mathbf{h}$ unique to the operator.
\begin{equation}
\begin{split}
L(\mathbf{u})
& = \mathbf{u} \ast \mathbf{h} \\
\end{split}
\end{equation}
\item
 Equivariance further demands that $\mathbf{h}$ be radially symmetric, eg separable into a scalar radial function $R(|\mathbf{r}|)$ and a unit tensor $\mathbf{Y_{l_{h}}}(\mathbf{\hat{r}})$.
 
    \begin{equation}
    \mathbf{h} = R \mathbf{Y_{l_{h}}}\\
    \end{equation}
    \begin{equation}
    \mathbf{Y_l(\mathbf{\hat{r}})}=
    \begin{cases}
    1 & \text{for $l=0$}\\
    \mathbf{\hat{r}} & \text{for $l=1$}\\
\text{higher order harmonics} & \text{for $l>1$}\\
    \end{cases}
\end{equation}
\item
 The scalar Fourier convolution theorem also extends to tensor fields where $\mathcal{F}$ is the component-wise tensor field Fourier transform 
$\mathbf{h} = R \mathbf{Y_{l_{h}}}\\$
    \begin{equation}
    \begin{split}
\mathcal{F}\{L(\mathbf{u})\}
& =\mathcal{F}\{ \mathbf{u} \ast \mathbf{h}\}\\
& =\mathcal{F}\{ \mathbf{u}\} \otimes \mathcal{F}\{ \mathbf{h}\}\\
& =\mathcal{F}\{ \mathbf{u}\} \otimes \mathcal{F}\{R \mathbf{Y_{l_{h}}}\}\\
& =\mathcal{F}\{R\} \mathcal{F}\{ \mathbf{u}\} \otimes \mathbf{Y_{l_{h}}}\\
    \end{split}
\end{equation}
\end{enumerate}
\end{theorem}
\end{mdframed}

Differential operators and Green's functions are in fact equivariant linear operators. For example, consider the operator mapping the scalar charge field to the vector electric field via Gauss's Law. The filter is a vector field symmetrically pointing outward with radial function $1/r^2$. The scalar vector product acts as multiplication. Other examples:


\begin{center}
\begin{tabular}{ |c|c|c|c| } 
 \hline
 Operator $L$  & $R(r)$ & $\mathbf{Y_{l_h}}(\mathbf{\hat{r}})$ & $(l_u,l_h)\rightarrow l_v$ \\
 \hline
 $I$ & $\delta(r)$ & 1 & $(l,0)\rightarrow l$ \\ 
 $\bigtriangledown$ & $\delta '(r)$ & $\mathbf{\hat{r}}$ & $(0,1)\rightarrow 1$  \\ 
 $\bigtriangledown \cdot$ & $\delta '(r)$ & $\mathbf{\hat{r}}$ & $(1,1)\rightarrow 0$ \\ 
 $\bigtriangledown \times$ & $\delta '(r)$ & $\mathbf{\hat{r}}$ & $(1,1)\rightarrow 1$ \\ 
  Diffusion process & Gaussian & 1 & $(0,0)\rightarrow 0$ \\
   Electric potential "$(\bigtriangledown^2)^{-1}$" & $ 1/r$ & 1 & $(0,0)\rightarrow 0$ \\ 
  Gauss's law "$(\bigtriangledown \cdot)^{-1}$" & $1/r^2$ & $\mathbf{\hat{r}}$ & $(0,1)\rightarrow 1$ \\ 
 \hline
\end{tabular}
\end{center}

\subsection{Neural equivariant operators for machine learning}
By changing the radial function, we change the filter kernel and thus the operator's behavior. This is useful in machine learning equivariant transformations between scalar or vector fields. Here we introduce neural equivariant operators 
$L_{\mathbf{p}}$.

\begin{mdframed}
\begin{definition}[Neural equivariant operators]
A neural equivariant operator is a linear equivariant operator $L_{\mathbf{p}}$ whose radial function $R(r;\mathbf{p})$ is parameterized with trainable parameters $\bold{p}$
\end{definition}

\end{mdframed}

\section{Software}

Our package EquivariantOperators.jl implements in Julia fully differentiable finite difference operators on scalar or vector fields in 2d/3d. It can run forwards for PDE simulation or image processing, or back propagated for machine learning or inverse problems. Emphasis is on symmetry preserving rotation equivariant operators, including differential operators, common Green's functions \& parametrized neural operators. Supports possibly nonuniform, nonorthogonal or periodic grids.

The package's full differentiability means the derivatives of any code (eg loss functions) using our operators can be computed automagically via automatic differentiation in Julia. This allows arbitrary objective or loss functions to be optimized via back propagation and gradient descent. Differentiating against radial function parameters of neural equivariant operators enables machine learning or deconvolution between scalar or vector fields. Differentiating against equation parameters enables parameter estimation in PDEs. Differentiating against the input field enables geometry optimization or solving inverse problems. Differentiating against control parameters or forcing functions enables solving optimal control problems.

\section{Experiments and discussion}
Tutorials for experiments hosted on Google Colab: \url{https://colab.research.google.com/drive/17JZEdK6aALxvn0JPBJEHGeK2nO1hPnhQ?usp=sharing} Star the package on Github if you found it useful.

\subsection{Equivariant machine learning on vector fields}
We learn the mappings from electric charge to electric potential and electric field. Both transforms are linear and equivariant so only one training sample suffices to train the two neural equivariant operators! Indeed, after training on an electric dipole configuration, the operators learn the $1/r$ and $1/r^2$ radial functions which completely characterize the dynamics. Unsurprisingly, this one shot model achieves low error on a test set of different electric charge densities.

\begin{figure}[H]

\begin{subfigure}{0.5\textwidth}
\includegraphics[width=0.9\linewidth, height=6cm]{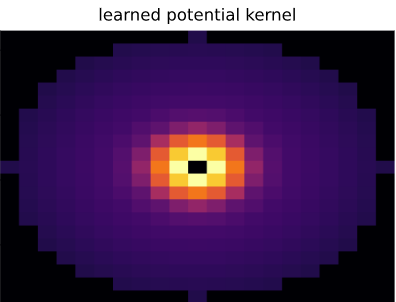} 
\label{fig:subim1}
\end{subfigure}
\begin{subfigure}{0.5\textwidth}
\includegraphics[width=0.9\linewidth, height=6cm]{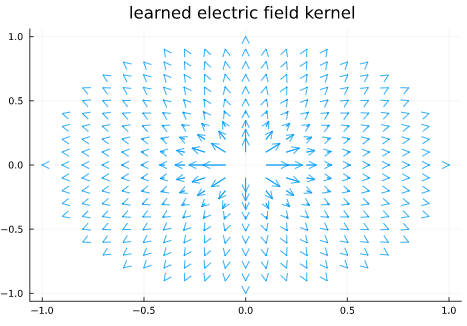}
\label{fig:subim2}
\end{subfigure}

\label{fig:image2}
\end{figure}

\subsection{Machine learning spatial temporal dynamics}
It's common to seek to learn spatial temporal dynamics from a "movie" of the system. In climate science, for example, one may wish to predict wild fire propagation, vegetation growth or cloud cover based on satellite images over time. In these problems, the approximate dynamics are known but empirical parameters are not. Often one can formulate smoothed partial differential equations consisting of empirically parameterized diffusion, advection and reaction. From an experimental movie, one can make a training set mapping the system states to their approximated time derivatives. We generate training data by simulating a diffusion advection PDE of a point emitter in a wind field. Training results in good fit of the parameters of diffusion coefficient and wind velocity.

\printbibliography 

\end{document}